\documentclass[letterpaper, 10 pt, conference]{ieeeconf}  

\IEEEoverridecommandlockouts                              

\usepackage{amsmath} 
\usepackage[dvipsnames]{xcolor}
\usepackage{booktabs}
\usepackage{multirow}
\usepackage{bm}
\usepackage{graphicx}
\usepackage{mathtools}
\usepackage{float}
\usepackage{subfigure}
\usepackage{overpic}
\usepackage{lipsum}
\usepackage{tabularx}
\title{\LARGE \bf
MOISST: Multimodal Optimization of Implicit Scene for SpatioTemporal calibration
}

\author{Quentin Herau$^{1,2}$,
Nathan Piasco$^{1}$,
Moussab Bennehar$^{1}$,
Luis Roldão$^{1}$,
Dzmitry Tsishkou$^{1}$,\\
Cyrille Migniot$^{3}$,
Pascal Vasseur$^{4}$
and Cédric Demonceaux$^{2}$
\thanks{$^{1}$ Noah's Ark, Huawei Paris Research Center, France. {\tt\small \{Quentin.Herau, Nathan.Piasco, Moussab.Bennehar, Luis.Roldao, Dzmitry.Tsishkou\}@huawei.com}}
\thanks{$^{2}$ ICB UMR CNRS 6303, Universit\'{e} de Bourgogne, France. {\tt\small \{Quentin.Herau@etu., Cedric.Demonceaux@\}u-bourgogne.fr }}
\thanks{$^{3}$ ImViA UR 7535, Universit\'{e} de Bourgogne, France. {\tt\small  Cyrille.Migniot@u-bourgogne.fr }}
\thanks{$^{4}$ MIS UR 4290, Universit\'{e} de Picardie Jules Verne, France. {\tt\small Pascal.Vasseur@u-picardie.fr}}
}

\begin{document}

\maketitle

\thispagestyle{empty}
\pagestyle{empty}


\begin{abstract}

With the recent advances in autonomous driving and the decreasing cost of LiDARs, the use of multimodal sensor systems is on the rise. However, in order to make use of the information provided by a variety of complimentary sensors, it is necessary to accurately calibrate them. We take advantage of recent advances in computer graphics and implicit volumetric scene representation to tackle the problem of multi-sensor spatial and temporal calibration. Thanks to a new formulation of the Neural Radiance Field (NeRF) optimization, we are able to jointly optimize calibration parameters along with scene representation based on radiometric and geometric measurements. Our method enables accurate and robust calibration from data captured in uncontrolled and unstructured urban environments, making our solution more scalable than existing calibration solutions. We demonstrate the accuracy and robustness of our method in urban scenes typically encountered in autonomous driving scenarios.
\end{abstract}

\section{INTRODUCTION}

Most robotic and intelligent systems rely heavily on sensory information to achieve various tasks.
Moreover, commonly encountered sensor setups for autonomous driving consist of multiple sensors acquiring different data modalities (e.g. cameras, LiDARs, IMUs, GNSS systems, etc.) which can greatly improve the performance on different tasks such as mapping~\cite{choi2012environment}, localization~\cite{vivacqua2017self} and perception~\cite{yoo20203d}.

However, to correctly exploit and merge the information provided by all sensors, it is important to represent their data in a common reference frame.
Spatial extrinsic calibration is the process that determines the relative geometric transformation between the sensor poses by considering a 6-DoF rigid-body transformation. Although accurate spatial calibration is essential in multi-sensor setups, it is often not sufficient due to time synchronization issues between the different sensors.
Time synchronization is the process that determines the time offset between the different sensor measurements, in the case there is no hardware synchronization which would eliminate any delay. 

Current existing methods commonly require the use of calibration targets placed in the scene to fuse all sensors in a common frame~\cite{zhang2004extrinsic,geiger2012automatic}. This is unpractical in many cases, especially for dynamic tasks, where the calibration setup must be redone regularly. Although some papers offer solutions to bypass this constraint by detecting salient geometric features (e.g. edges~\cite{napier2013cross,yuan2021pixel}, planes~\cite{rehder2014spatio}) within acquired scenes, such features might not be present in all kind of environments. 

\begin{figure}
     \centering
      \begin{subfigure}
         \centering
         \includegraphics[width=.45\columnwidth]{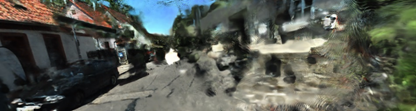}
     \end{subfigure}
     \hfill
     \begin{subfigure}
         \centering
         \includegraphics[width=.45\columnwidth]{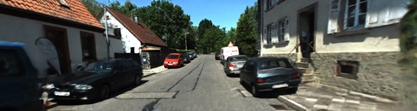}
     \end{subfigure}
     \begin{subfigure}
         \centering
         \includegraphics[width=.45\columnwidth]{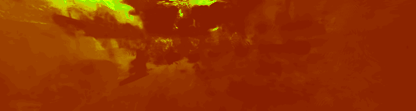}
     \end{subfigure}
     \hfill
     \begin{subfigure}
         \centering
         \includegraphics[width=.45\columnwidth]{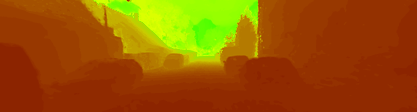}
     \end{subfigure}
     \begin{subfigure}
         \centering
         \includegraphics[width=.45\columnwidth]{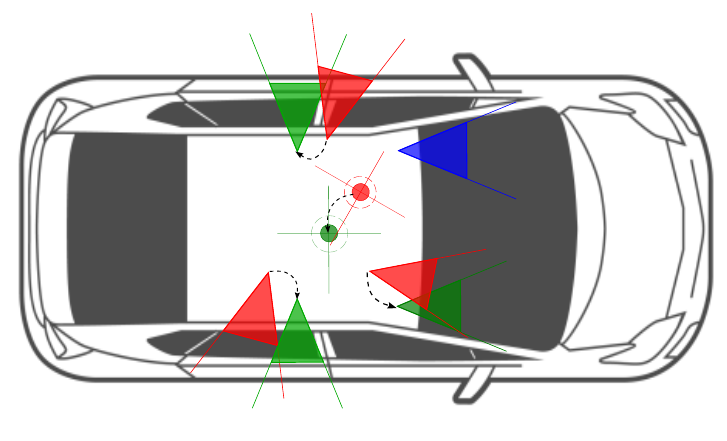}
     \end{subfigure}
     \hfill
     \begin{subfigure}
         \centering
         \includegraphics[width=.45\columnwidth]{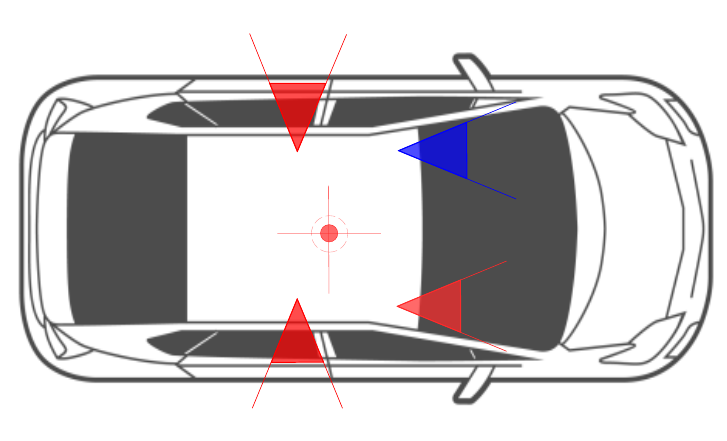}
     \end{subfigure}
     \caption{Effect of calibration on novel view synthesis: training positions in \textcolor{red}{red}, ground truth positions in  \textcolor{ForestGreen}{green}, reference position in \textcolor{blue}{blue}. The RGB images (top) and depth maps (middle) are rendered from an implicit neural 3D scene trained from non-calibrated (left) and calibrated with MOISST (right) sensors.
     }
    \label{fig:header}
\end{figure}

Furthermore, these methods often do not consider the possible asynchronization of the sensors. The effect of a wrongly synchronized rig can have a consequential impact on the performance depending on the task. Moreover, unconsidered time offsets within calibration can strongly degenerate extrinsic estimation leading to suboptimal results. 

Considering all the issues mentioned above, we introduce our new method called MOISST: Multimodal Optimization of Implicit Scene for SpatioTemporal calibration. MOISST is a novel calibration method which leverages an implicit neural 3D scene representation known as Neural Radiance Fields (NeRF)~\cite{mildenhall2021nerf}. It can be trained with any kind of sensor providing radiometric or geometric information on a given 3D scene. This representation is by nature the common reference frame used for the sensor fusion. We take advantage of the differentiable property of our scene representation to simultaneously learn the scene's geometry and colors, and the poses given to the neural network. Unlike existing NeRF-based methods of pose regression~\cite{wang2021nerf,jeong2021self,lin2021barf}, we consider the rigid constraint in the multi-sensor rig to reduce the number of optimized parameters. By using a time-parameterized differentiable formulation for the main sensor trajectory, we can also detect and compensate potential time offsets between the sensors. To the best of our knowledge, tackling the problem of multi-sensor spatiotemporal calibration using implicit representations has not been proposed before in the literature.

Thanks to our formulation, we are able to propose an offline targetless solution to spatiotemporal calibration of multimodal sensors, that is also structureless, as we do not require specific geometric structures like edges or planes in the scene for our method. Compared to other methods and because of the aforementioned characteristics of our solution, MOISST is especially adapted to perform automatic re-calibration of a multi-sensor device during the full life-cycle of the system. MOISST is a simple -- \textit{it can be run from acquisition data recorded in any environment} -- and inexpensive -- \textit{it does not require target or external hardware} -- calibration solution, which are crucial features for robots and large scale fleet of autonomous vehicles.

\section{RELATED WORK}\label{sec:related_work}

\subsection{Multimodal extrinsic and temporal calibration}\label{sec:calibration}

Extrinsic calibration for multimodal sensors is a well studied subject that can be categorized in two main groups: target-based and targetless methods.

\subsubsection{Target-based calibration}
Zhang \textit{et al.}~\cite{zhang2004extrinsic} were the first to introduce the use of a planar checkerboard target for camera and laser range finder calibration, by using the latter to determine the checkerboard plane, and the pattern seen by the camera to calculate its pose. Three pairs of capture measurements are enough to deduce extrinsic parameters between both sensors.
Geiger \textit{et al.}~\cite{geiger2012automatic} propose a solution to calibrate the sensors with a single capture, by placing multiple targets in the scene. While these methods provide satisfactory calibration accuracy, they necessitate the placement of targets in the scene, which might not be available or practical to place in typical real-world scenarios.

\subsubsection{Targetless calibration}
Targetless methods usually use the concept of mutual information, by matching corresponding elements obtained through different type of sensors. This can be edges for visible cameras, depth gradient for LiDARs~\cite{napier2013cross,yuan2021pixel}, or the use of correspondence between image intensity and surface normals~\cite{taylor2012mutual}. However, by relying on specific geometric features, these methods only work in a well structured scene with noticeable, recognizable and detectable patterns such as straight lines or edges and often work only in indoor environments. There is also a set of deep learning based methods~\cite{schneider2017regnet,iyer2018calibnet,lv2021lccnet}, able to find the transformation between a camera capture and a LiDAR scan. However, these methods are trained in a supervised manner, needing a labeled dataset, and are prone to overfitting, limiting their use to environments reflecting the training dataset.

Although the previously mentioned methods achieve satisfactory performance given ideal conditions, they suppose a perfectly time-synchronized set of sensors, which is possible through specific hardware~\cite{sommer2017low} but is often challenging and sensor dependant (i.e. most low-cost cameras do not support such features). There exists some methods tackling temporal calibration, both target-based~\cite{kelly2014general}, or targetless~\cite{rehder2014spatio}, they require however additional sensors such as calibrated camera-IMU pair to obtain a precise trajectory. With the methods proposed by Taylor \textit{et al.}~\cite{taylor2016motion} and Park \textit{et al.}~\cite{park2020spatiotemporal}, visual and LiDAR odometry are used for calculating trajectories for each sensor and matching them,  allowing both spatial and temporal calibration. Nevertheless, this approach can generate a progressive drift with the accumulated transformations between the frames.

Contrary to the previously mentioned methods, MOISST does not require any targets (i.e. targetless) and determines both spatial and temporal calibration parameters by only relying on the poses of a single sensor, avoiding the cumulative errors in the different per-sensor trajectories. At the same time, we fuse information from all sensors, and thanks to the use of a dense implicit scene representation, we do not require specific geometric structure (i.e. structureless). This approach allows compatibility with a greater variety of scenes and allows us to scale to almost all real-world scenarios.

\subsection{Neural 3D scene representation}
NeRF~\cite{mildenhall2021nerf} is an implicit representation of a 3D scene primarily used for novel view synthesis. From a set of images and their corresponding camera poses, the model learns the 3D geometry through differentiable volume rendering. NeRF provides a continuous representation, resulting in improved rendering fidelity and compactness compared to classical explicit scene representations~\cite{Waechter2014LetTB}. Beyond the rendering ability, many recent methods have used these implicit scene representations for downstream robotics tasks~\cite{imap, IchnowskiAvigal2021DexNeRF, lens}. 

NeRF stores all the color and density information of the scene in a multilayer perceptron (MLP), and allows any rendering resolution as the representation is continuous. The model takes as input a 3D coordinate and a direction vector, outputs a color and density information for this 3D point, and is trained through a differentiable rendering procedure. A sinusoidal encoding~\cite{tancik2020fourier} of the input coordinate maps the low dimensional 3D position and direction to a higher dimension representation, allowing the rendering of a highly detailed scene representation.
To speed-up convergence, Instant Neural Graphics Primitives (Instant-NGP)~\cite{muller2022instant} was introduced, allowing much faster convergence with higher quality rendering. It uses a multi-resolution hash encoding instead of sinusoidal encoding, considerably reducing the size of the trained MLP.

The training of NeRF requires mainly RGB images from cameras, optionally depth information such as point clouds from LiDARs~\cite{rematas2022urban}, along with registered poses for each sensor frame. The final rendering quality is highly dependant on the precision of these poses, as seen in Fig.~\ref{fig:header}, where the result without optimization has incorrect geometry, produces low quality novel views and is not usable. As an answer to this limitation, NeRF$^{--}$~\cite{wang2021nerf} exploits the fully differentiable structure of NeRF to not only train the NeRF model, but also to optimize the camera poses. This makes the model robust to noisy poses, as it is able to optimize both the camera poses alongside the NeRF model. SCNeRF~\cite{jeong2021self} uses the first two columns of the rotation matrix to formulate rotations instead of Rodrigues formula to achieve better convergence, and BARF~\cite{lin2021barf} improves upon these methods by using low-to-high frequency release for the input positional encoding, avoiding local minima during pose optimization.

In our proposal, because we focus on multi-sensor calibration, we only optimize the extrinsic transformation between the sensors instead of optimizing each pose of each frame independently. Indeed, in a rigid sensor setup, sensors are not allowed to move freely relative to each other.
Compared to aforementioned methods, this novel and calibration-focused formulation reduces the number of parameters to be optimized and is more robust to outliers thanks to the rigidity constraint imposed on each pose. The proposed formulation also allow us to optimize the time offset between sensors, which is often hard to be achieved within the same optimization framework and may require additional information. 
    





    

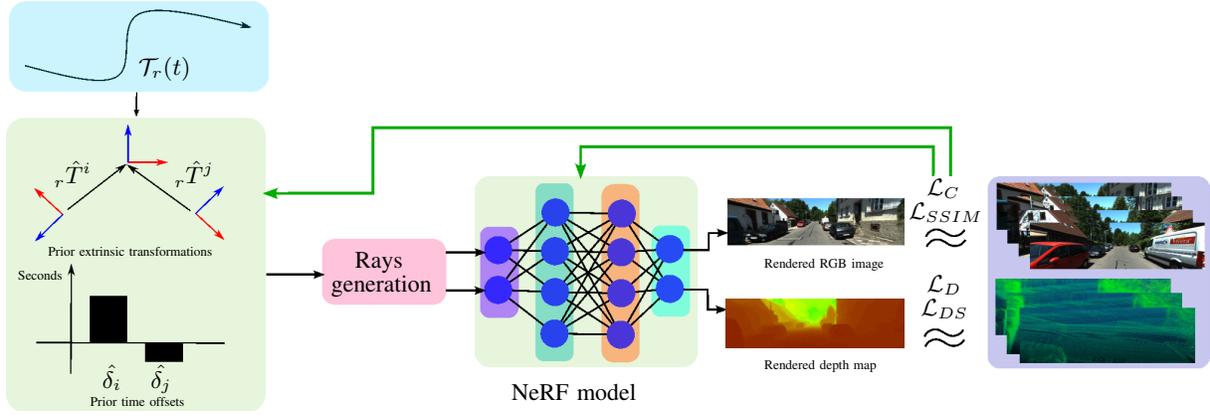
\begin{figure*}[ht]
    \centering
    \begin{overpic}[width=.9\textwidth]{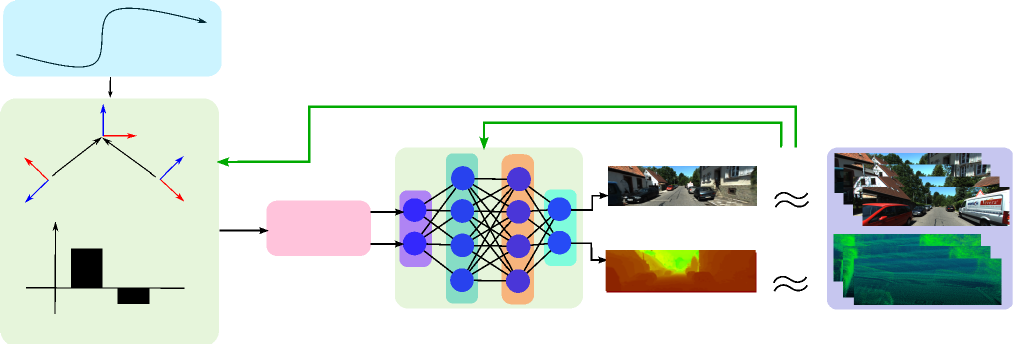} 

    \put(11, 28){\small $\mathcal{T}_r(t)$}

    \put(4, 19){\small $\prescript{}{r}{\hat{T}}^i$}
    \put(14, 19){\small $\prescript{}{r}{\hat{T}}^j$}
    \put(3.5, 13){\tiny Prior extrinsic transformations}

    \put(1, 11){\tiny Seconds}
    \put(7, 0.5){\tiny Prior time offsets}
    \put(8, 2){\small $\hat{\delta_i}$}
    \put(12, 2){\small $\hat{\delta_j}$} 

    \put(29, 12){\small Rays}
    \put(27, 10){\small generation}

    \put(76.5, 18){\small $\mathcal{L}_{C}$}
    \put(75, 16){\small $\mathcal{L}_{SSIM}$}

    \put(76.5, 10){\small $\mathcal{L}_{D}$}
    \put(76, 8){\small $\mathcal{L}_{DS}$}

    \put(42, 1){\small NeRF model}

    \put(63, 12){\tiny Rendered RGB image}
    \put(63, 3.5){\tiny Rendered depth map}
    
    \end{overpic}
    \caption{Overview of MOISST optimization framework. First, the model is initialized with rays generated using rough spatial and temporal calibration priors in addition to the reference frame trajectory. After each optimization step, the rays are regenerated and fed to the NeRF model. We then render RGB images and depth maps which are used along the ground truth ones to compute the losses and propagate the gradients. Gradient descent algorithm is finally used to optimize both NeRF and calibration parameters.} 
    \label{fig:framework2}
\end{figure*}
\section{METHOD}

\subsection{Notations and background}

\subsubsection{Notations}

We consider a multi-sensor system with $S$ sensors with $r \in [1, S]$ being our reference sensor and each sensor is either a camera or a LiDAR. We use the following notations to describe our method:
\begin{itemize}
    \item $ {\{N_i\}}_{i \in [1, S]}$ : set of number of frames captured by each sensor,
    \item $ n_i, n \in [1,N_i] $: index of frames captured by sensor $i$,
    \item $t^{n_i} \in \mathbf{R^+} $: the absolute timestamp of frame $n_i$ relative to the sensor $i$ clock,
    \item $\delta_i \in \mathbf{R}$: the time offset between the reference frame clock and the sensor $i$ clock,
    \item $\prescript{}{w}{T}^i(t)\in \mathbf{R}^{4\times 4}$: the pose transformation matrix of sensor $i$ at time $t$ (time relative to sensor $i$ clock) in the world reference, 
    \item $\prescript{}{j}{T}^i \in \mathbf{R}^{4 \times 4}$: the extrinsic homogeneous transformation matrix from sensor $i$ to sensor $j$.
\end{itemize}

We aim to calibrate our system according to the reference sensor. The goal is to obtain the transformation matrices  $\prescript{}{j}{T}^i$ and time offsets $\delta_i$ between the sensors to calibrate and the reference one. We consider that we know the pose of sensor $r$ in a global frame, which could be easily obtained through SLAM~\cite{mur2015orb} or Structure-from-Motion~\cite{colmap}. We also consider its clock as the reference clock, $\delta_r=0$. From poses of reference sensor, $\prescript{}{w}{T}^r(t^{n_r}), n_r \in [1, N_r]$, we build a continuous trajectory. We do that by interpolating between the existing poses, and extrapolating outside the defined temporal bounds by extending the transformations at the beginning and the end of the sequence. This modeling process is very similar to what is defined in~\cite{park2020spatiotemporal}.
For the interpolation functions, we used spherical linear interpolation (SLERP)~\cite{shoemake1985animating} for the rotation, and linear interpolation (LERP) for the translation. We denote this interpolation function as $\mathcal{T}_r$:
\begin{equation}
    \prescript{}{w}{T}^r(t) = \mathcal{T}_r(t).
\end{equation}

\subsubsection{Spatiotemporal calibration}

Given the spatial extrinsic calibration and the time offset of the other sensors regarding our reference sensor, we can compute the pose of sensor $i$ in a global frame with the formula:
\begin{equation}
     \prescript{}{w}{T}^i(t^{n_i}+\delta_{i}) =\mathcal{T}_r(t^{n_i}+\delta_{i}) \prescript{}{i}{T}^r.
     \label{eq:ext}
\end{equation}

\subsubsection{Implicit neural scene representation}

An implicit neural representation models a scene with a neural network by mapping coordinates as inputs to quantities of interests, such as color or density, as outputs. By evaluating points along camera rays and composing their densities and colors through volumetric rendering, such methods can synthesize RGB images and depth maps\footnote{We can estimate depth of ray with alpha composition of distances from the center of the ray to the sampled points.} from an arbitrary sensor pose.

In order to train said neural network on a specific scene, it is necessary to have a training set with sensor information and a pose associated. This information may be an RGB image in the case of visible camera, or a point cloud in the case of a LiDAR.
We aim to find the parameters of the neural network $\Theta$ that minimizes the difference between the provided information (${I}^{n_i}$ - image $n_i$ of sensor $i$ - or ${D}^{n_i}$ - depth information $n_i$ of sensor $i$) and the rendered result by the model defined as:
\begin{eqnarray}
\mathcal{R}_I\left({\prescript{}{w}{T}^i(t^{n_i})} \mid \Theta\right),\label{eq:render_function}\\
\mathcal{R}_D\left({\prescript{}{w}{T}^i(t^{n_i})} \mid \Theta\right),
\end{eqnarray}
with $\mathcal{R}_I$ being the model inference and ray composition function that returns a RGB image prediction of frame $n_i$ for sensor $i$ and $\mathcal{R}_D$ being equivalent to $\mathcal{R}_I$ but returning rays depth instead of colors.
By minimising the loss $\mathcal{L}_{total}$ defined as:
\begin{eqnarray}
\mathcal{L}_{total} &=& \lambda_C \mathcal{L}_{C}+\lambda_D \mathcal{L}_{D}, \label{eq:losses}\\
\mathcal{L}_{C} &=& \sum_{i=1}^{S} \sum_{n_i=1}^{N_i}\left\|\mathcal{R}_I\left({\prescript{}{w}{T}^i(t^{n_i})} \mid \Theta\right)-{I}^{n_i}\right\|_{2}^{2}, \label{eq:color_loss}\\
\mathcal{L}_{D} &=& \sum_{i=1}^{S} \sum_{n_i=1}^{N_i}\left\|\mathcal{R}_D\left({\prescript{}{w}{T}^i(t^{n_i})} \mid \Theta\right)-{D}^{n_i}\right\|_{2}^{2}, \label{eq:depth_loss}
\end{eqnarray}
with $\lambda_C$, $\lambda_D$ weighting hyper-parameters, we can estimate the optimal network parameters $\hat{\Theta}$ satisfying:
\begin{equation}
    \hat{\Theta}=\underset{\Theta}{\text{argmin}}(\mathcal{L}_{total}).
\end{equation}

As explained by Wang \textit{et al.}~\cite{wang2021nerf}, because the scene representation we use is fully differentiable, it is possible to optimize the input poses with gradient descent jointly with the radiance field parameters. The optimization objective becomes the following:
\begin{equation}
    \left\{ \hat{\Theta},\prescript{}{w}{\hat{T}}^i \right\} = \underset{\Theta, \prescript{}{w}{T}^i}{\text{argmin}}(\mathcal{L}_{total}). \label{eq:opt_barf}
\end{equation}



\subsection{MOISST Optimization formulation}

In this section, we introduce our novel optimization formulation for multi-sensor system spatiotemporal calibration.
Considering a multi-sensor system such as a robot or an autonomous car, we know that the poses of each sensor observation are not independent, as there exists a rigid transformation between each sensor. Because we know the trajectory $\mathcal{T}_r$ of the reference sensor $r$, we can express the absolute pose of each remaining sensor according to sensor $r$ (see equation~\ref{eq:ext}). Substituting $\prescript{}{w}{T}^i$ in equation~\ref{eq:render_function} leads to the following formulation:
\begin{equation}
    \mathcal{R}_I\left(\mathcal{T}_r(t^{n_i}+\delta_{i}) \prescript{}{i}{T}^r  \mid \Theta\right).
    \label{eq:render_w_ext}
\end{equation}
Similar reasoning can be made for depth rendering function $\mathcal{R}_D$. Our new formulation of the rendering functions can be integrated in the color loss of equation~\ref{eq:color_loss}:
\begin{equation}
    \mathcal{L}_{C} = \sum_{i=1}^{S} \sum_{n_i=1}^{N_i}\left\| \mathcal{R}_I\left(\mathcal{T}_r(t^{n_i}+\delta_{i}) \prescript{}{i}{T}^r  \mid \Theta\right)-{I}^{n_i}\right\|_{2}^{2}.
\end{equation}
We can replace the depth rendering function in the same manner in equation~\ref{eq:depth_loss}. This leads to our new optimization formulation:
\begin{equation}
    \left\{ \hat{\Theta},\prescript{}{i}{\hat{T}}^r,\hat{\delta_i} \right\} =\underset{\Theta,\prescript{}{i}{T}^r,\delta_i}{\text{argmin}}(\mathcal{L}_{total}),
\end{equation}
with $\prescript{}{i}{\hat{T}}^r\text{ and }\hat{\delta_i}$ the only parameters to optimize along with the network weights. As the trajectory $\mathcal{T}_r$ is continuous over time, we can also optimize the time offsets $\delta_i$. With the proposed method, we only have to optimize the extrinsic transformation between all sensors and the reference sensor, reducing the number of optimized parameters compared to the full set of frame poses as in equation~\ref{eq:opt_barf}.
A summary of our proposal is shown in Fig.~\ref{fig:framework2}.

\begin{center}
\begin{table*}[hbt!]
\centering
\scriptsize
\caption{\label{tab:spatial_calibration} Spatial calibration accuracy.}
\renewcommand{\arraystretch}{1.2}
\begin{tabular}{@{} l l ll l ll l l @{}}
\toprule
& \multicolumn{2}{c}{Front-right camera} && \multicolumn{2}{c}{LiDAR} \\
\cmidrule{2-3} \cmidrule{5-6}
Sequence & Translation error (cm) & Rotation error (°) && Translation error (cm) & Rotation error (°) \\
\hline
0 & $2.3\pm0.6$ & $0.09\pm0.02$ && $8.2\pm1.8$ & $0.21\pm0.08$ \\
1 & $1.6\pm0.4$ & $0.09\pm0.02$ && $7.3\pm2.1$ & $0.41\pm0.08$ \\
2 & $1.2\pm0.6$ & $0.04\pm0.03$ && $13.5\pm1.8$ & $0.18\pm0.10$ \\ 
4 & $2.2\pm0.7$ & $0.08\pm0.06$ && $9.9\pm2.6$ & $0.30\pm0.10$ \\ 
\bottomrule
\end{tabular}
\end{table*}
\end{center}
\subsection{Optimization details}

\subsubsection{Additional losses for geometric consistency}

We add two more losses using image patches to further improve the geometry of the NeRF model and the proper estimation of calibration parameters. The first loss is the structural dissimilarity (DSSIM) $\mathcal{L}_{SSIM}$~\cite{ssim}, which minimizes the difference in local 2D structures between the rendered and the input image~\cite{wang2023planerf}. The second is the depth smoothness loss $\mathcal{L}_{DS}$, also used by RegNeRF~\cite{niemeyer2022regnerf}, which regularizes the depth variation in randomly selected patches of the images to reduce variation of the predicted depth. Our final loss function becomes as follows:
\begin{equation}
    \mathcal{L}_{total}=\lambda_C \mathcal{L}_{C}+\lambda_D \mathcal{L}_{D}+\lambda_{SSIM} \mathcal{L}_{SSIM}+\lambda_{DS} \mathcal{L}_{DS}
\end{equation}
with $\lambda_C,\lambda_D,\lambda_{SSIM},\lambda_{DS}$ the weight factor for each loss.
\subsubsection{Network architecture}
We use an implicit scene representation similar to the \texttt{nerfacto} model of Nerfstudio\footnote{https://docs.nerf.studio/en/latest/nerfology/methods/nerfacto.html} open source framework. It is inspired by the proposal network introduced in MipNeRF 360~\cite{barron2022mip} with two proposal radiance fields and one final radiance field that outputs the color and density for the volumetric rendering. The proposals are used to samples points along the rays where the density is high. We use the hash grid introduced in instant NGP~\cite{muller2022instant} for positional encoding and spherical harmonics for directional encoding. We found this model to be a good trade-off between speed and accuracy for our spatiotemporal calibration problem.

\subsubsection{Regularization}
\label{sec:reg}
In BARF~\cite{lin2021barf}, the idea of low-to-high frequency release for the positional encoding allows smoothness in the scene, which helps the optimization of the poses to avoid local minima. 
With our architecture using the instant-NGP backbone, we do not have a sinusoidal positional encoding, but a multi-resolution hash grid instead. In order to mimic BARF, we introduce a weight decay to the hash encoder for a few epochs, before removing it, allowing higher frequency information to be learned afterwards.
We also wait a few epochs before applying the depth loss $L_D$, as we found that initializing the geometry solely through visual supervision helps the whole system to converge better.

\subsubsection{Spatiotemporal priors}
We use spatial calibration priors to initialize the extrinsic parameters of the sensors. For the temporal shift, we set the initial estimate to 0 as we found that our solution was very robust to temporal calibration. Ablation on the sensibility of our method to initial prior is provided in section~\ref{sec:abla}.

\section{EXPERIMENTS}\label{sec:experiments}

We evaluate MOISST on the NVS training set from the recent KITTI-360 dataset~\cite{liao2022kitti} which involves difficult static outdoor scenarios with 2 forward and 2 side facing cameras and a top-mounted Velodyne HDL-64E LiDAR sensor. The Lidar scans provided by the KITTI-360 dataset are undistorted using motion compensation. We report results on sequences 0, 1, 2 and 4 and consider the front-left camera as our reference sensor $r$ for all experiments\footnote{Experiments are not performed on Sequence 3 as it has missing captures of LiDAR scans.}. We apply $\pm50$ cm translation and $\pm5^\circ$ rotation offsets on all axes and $\pm100$ ms time offset to simulate spatial and temporal calibration errors, respectively. The geodesic distance and the classical $L_2$-norm are used to report the rotation and translation errors along all axes, and the average error over the last 10 epochs is reported for fairness. Each training takes between 1 and 4 hours depending on the number of selected sensors.

\begin{center}
\begin{table}[b]
\centering
\scriptsize
\caption{\label{tab:temporal_ablation} Temporal calibration accuracy.}

\renewcommand{\arraystretch}{1.2}
\begin{tabular}{@{} l l ll l ll l l @{}}
\toprule
&& Front-right camera && LiDAR \\
\cmidrule{3-3} \cmidrule{5-5}
Sequence & Initial error (ms) & Temporal error (ms) && Temporal error (ms) \\
\hline
\multirow{2}{*}{0}
& 100  & $0.7\pm0.3$ && $7.6\pm1.0$ \\
& 200  & $0.5\pm0.3$ && $6.2\pm2.0$ \\
\hline
\multirow{2}{*}{1}
& 100  & $0.2\pm0.2$ && $1.5\pm1.1$ \\
& 200  & $0.5\pm0.3$ && $1.1\pm0.5$ \\
\hline
\multirow{2}{*}{2}
& 100  & $1.0\pm0.3$ && $14.0\pm1.8$ \\
& 200  & $1.1\pm0.3$ && $13.8\pm1.2$ \\
\hline
\multirow{2}{*}{4}
& 100  & $0.6\pm0.3$ && $3.4\pm1.5$  \\
& 200  & $0.7\pm0.3$ && $14.2\pm1.5$ \\
\bottomrule
\end{tabular}
\end{table}
\end{center}
\textbf{Implementation details:} \label{subsec:imp_detail} Given the sparser supervision signal from re-projected LiDAR depth maps, color and depth losses are balanced by ${\lambda_C=1}$ and ${\lambda_D=20}$. Furthermore, as radiance fields require initial optimization to learn depth through color supervision, depth loss is applied after two epochs. 
Geometric consistency losses are empirically balanced by ${\lambda_{SSIM}=0.1}$ and
${\lambda_{DS}=0.0001}$. We use the Adam optimizer and train our network during 50 epochs for all experiments.
We start with a learning rate of $1\times10^{-2}$ for the network parameters and $5\times10^{-5}$ for the spatial and temporal parameters, and exponentially decay to a factor of $1\times10^{-2}$ of the original learning rate.
We apply a weight decay of $1\times10^{-6}$ on the network parameters, including the hash grid, then remove the weight decay of the hash grid after 5 epochs as explained in section~\ref{sec:reg}.

We perform extensive evaluation of our solution upon three different conditions: in section~\ref{sec:spatial_only} with a scenario where we consider only spatial extrinsic noise, in section~\ref{sec:temp_only} with only temporal miscalibration and finally in section~\ref{Combined Calibration} by taking into account both spatial and temporal calibration parameters.

\subsection{Spatial Calibration}
\label{sec:spatial_only}
In this section, we only consider a spatial error, and remove the time offsets. We only optimize spatial parameters.
As we can see in Table~\ref{tab:spatial_calibration}, MOISST estimates calibration parameters with low rotation error, below 1° in all sequences. As for translation, we are able to reach around 2 cm of error on the front-right camera, and between 7 cm and 14 cm for the LiDAR. The higher error in LiDAR translation calibration needs to be mitigated because of the relative precision of the provided ground truth, as explained in section~\ref{sec:limitation_kitti}. 


\subsection{Temporal Calibration}
\label{sec:temp_only}
In this section, we only consider a temporal error, and remove the initial spatial error. We only optimize temporal parameters.
As shown by the results in Table~\ref{tab:temporal_ablation}, our method in able to temporally calibrate the front-right camera with a precision under 1 ms, despite the high initial time offset of 100 ms or 200 ms, knowing the camera is capturing at around 10 fps.
The LiDAR's final temporal error is more variable depending on the scene, reaching around 15 ms at maximum. As mentioned previously, this higher error might partially explain by the relative precision of the provided ground truth (see section~\ref{sec:limitation_kitti}).

\begin{figure*}
    \centering
    {\footnotesize
    \begin{tabularx}{17.5cm}{X X X X}
        \multicolumn{2}{c}{Camera pose (real image)} & \multicolumn{2}{c}{LiDAR pose (synthetic image)} \\
         KITTI extrinsic & Our extrinsic & KITTI extrinsic  & Our extrinsic 
    \end{tabularx}}
    
    \begin{subfigure}
        \centering
        \includegraphics[width=.5\columnwidth]{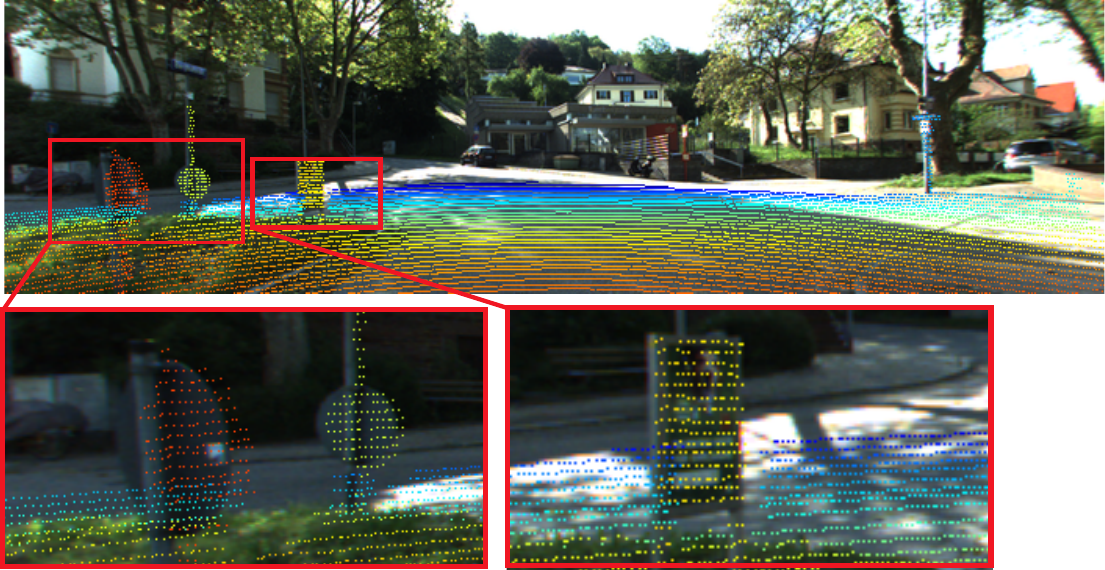}
        \includegraphics[width=.5\columnwidth]{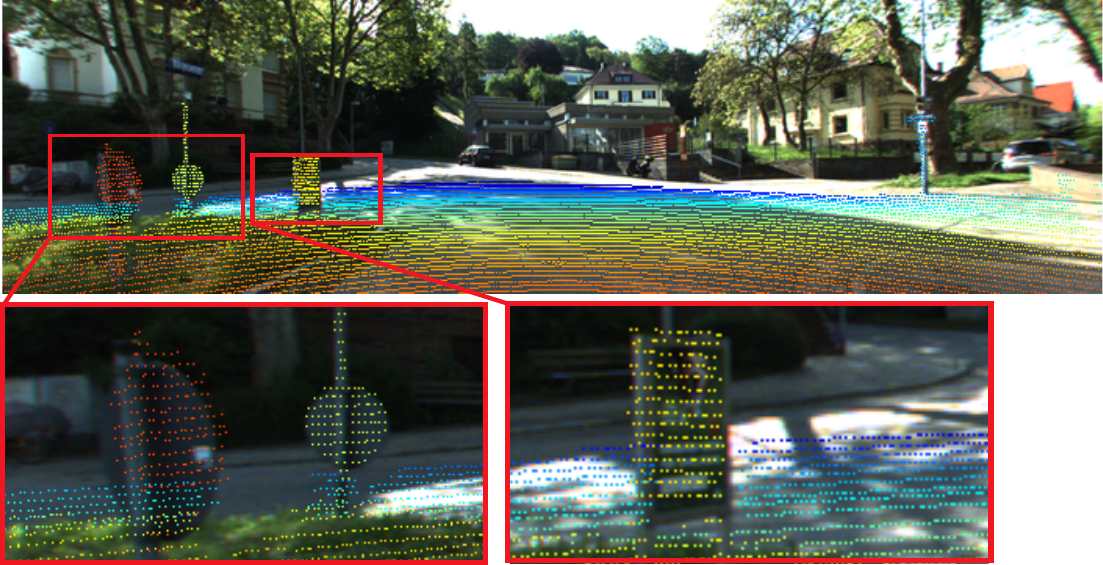}
    \end{subfigure}
    \hfill
     \begin{subfigure}
        \centering
        \includegraphics[width=.5\columnwidth]{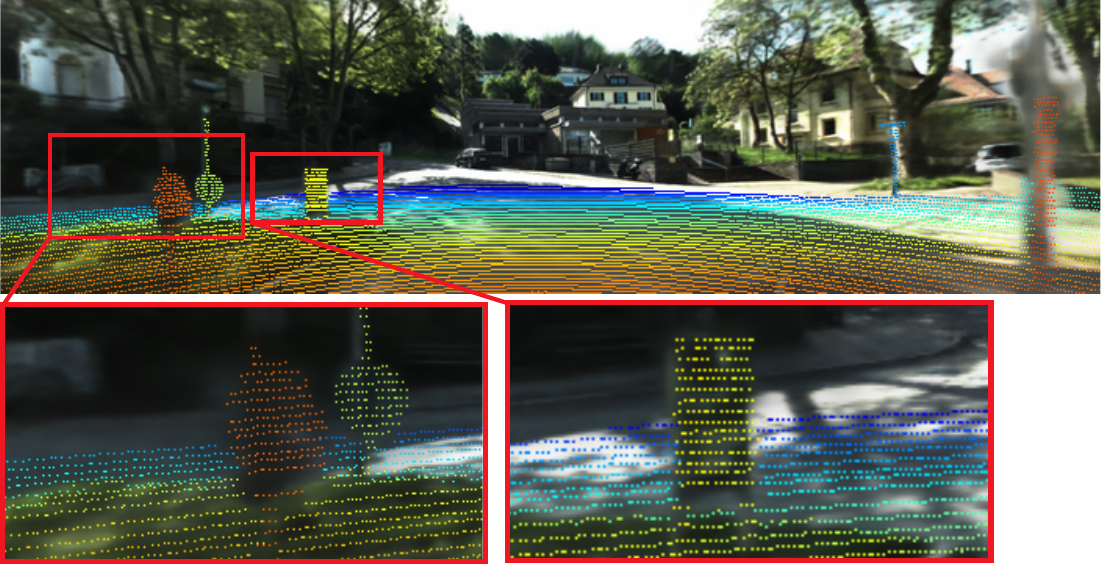}
        \includegraphics[width=.5\columnwidth]{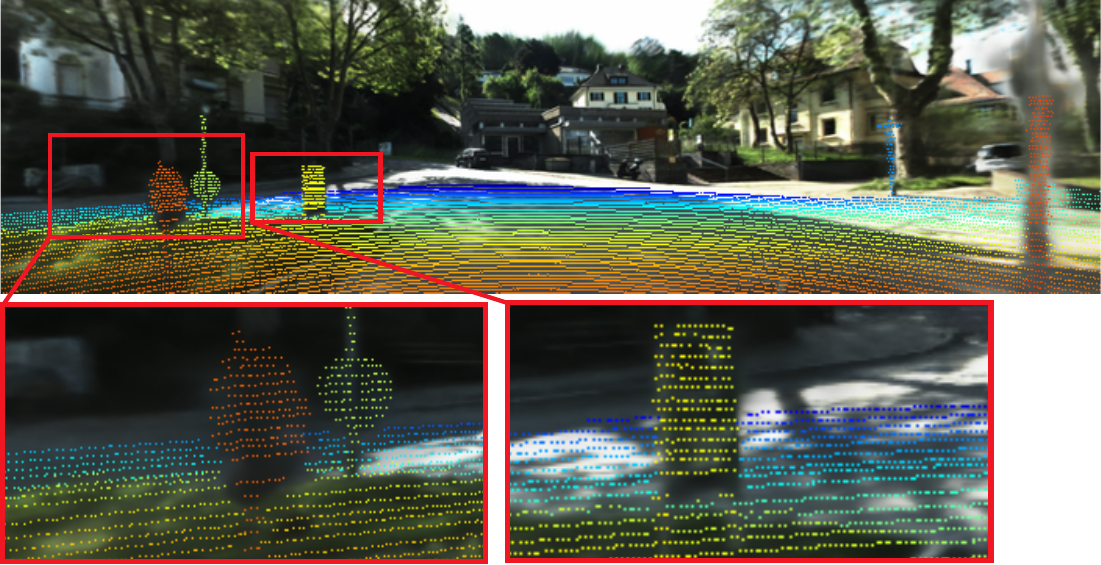}
    \end{subfigure}
    \begin{subfigure}
        \centering
        \includegraphics[width=.5\columnwidth]{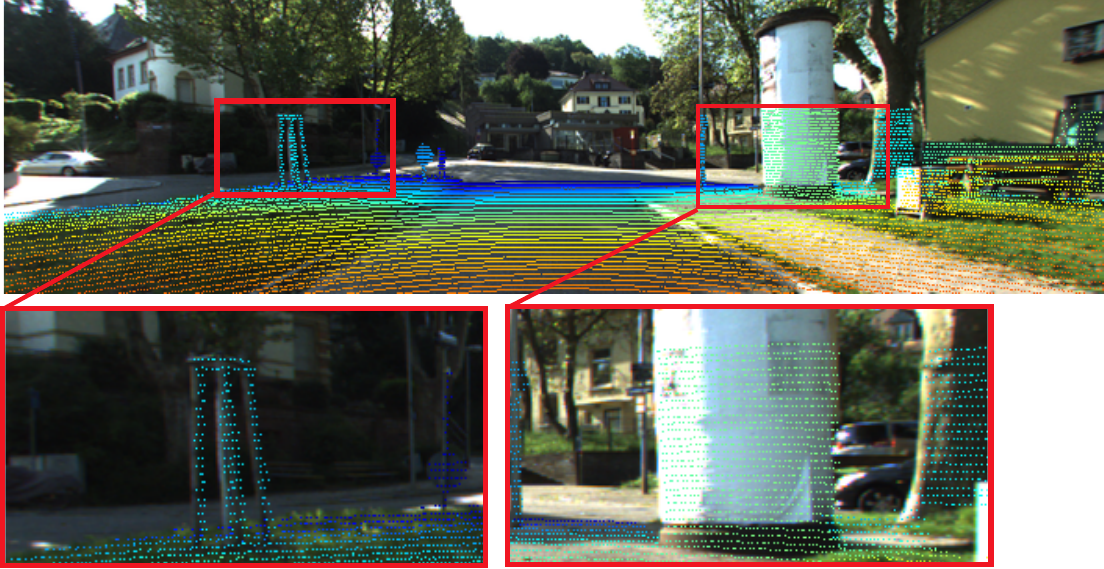}
        \includegraphics[width=.5\columnwidth]{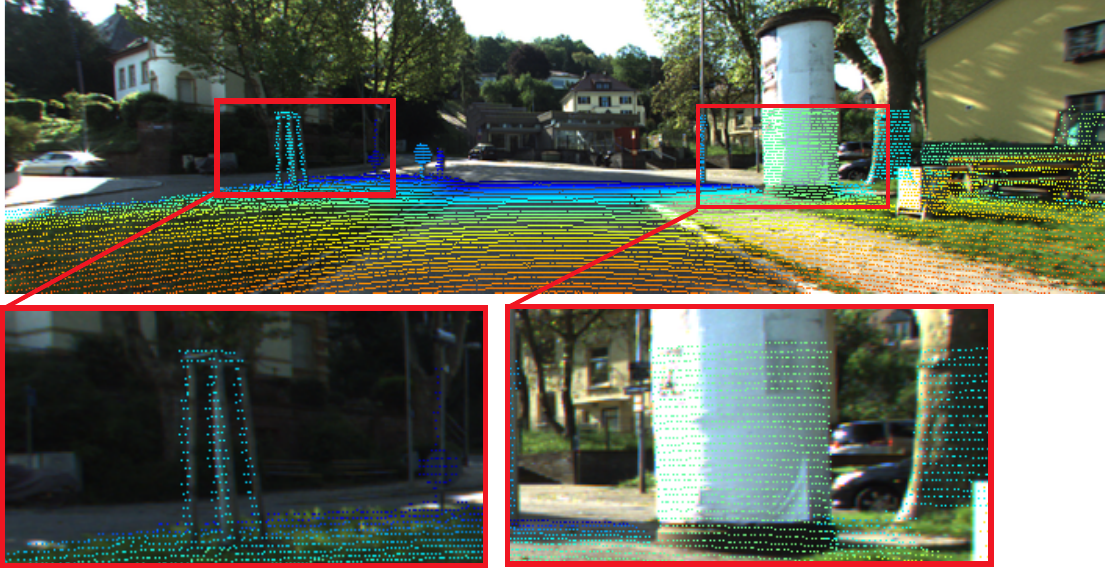}
    \end{subfigure}
    \hfill
     \begin{subfigure}
        \centering
        \includegraphics[width=.5\columnwidth]{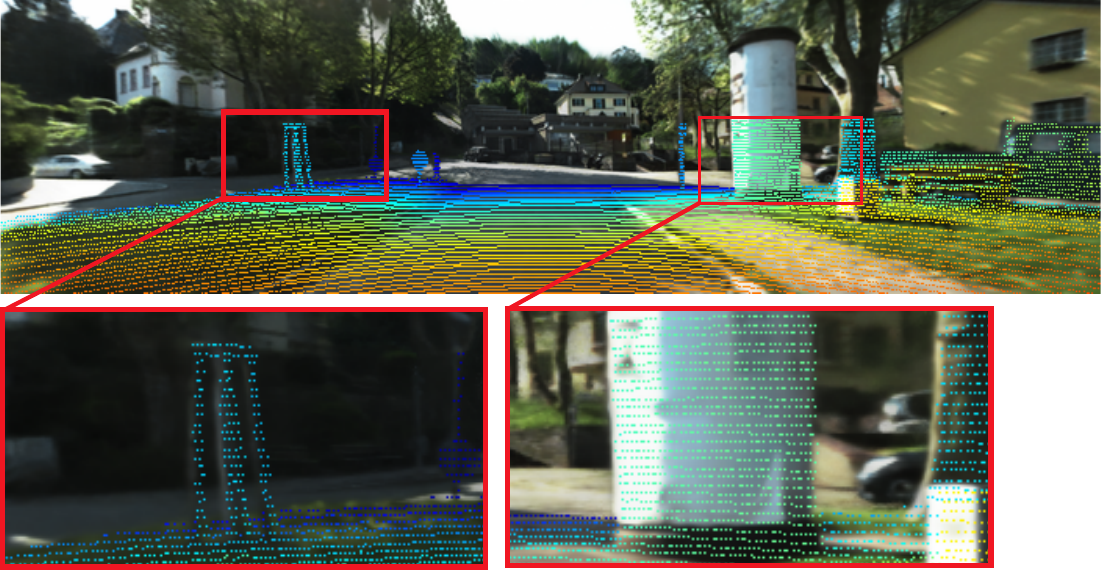}
        \includegraphics[width=.5\columnwidth]{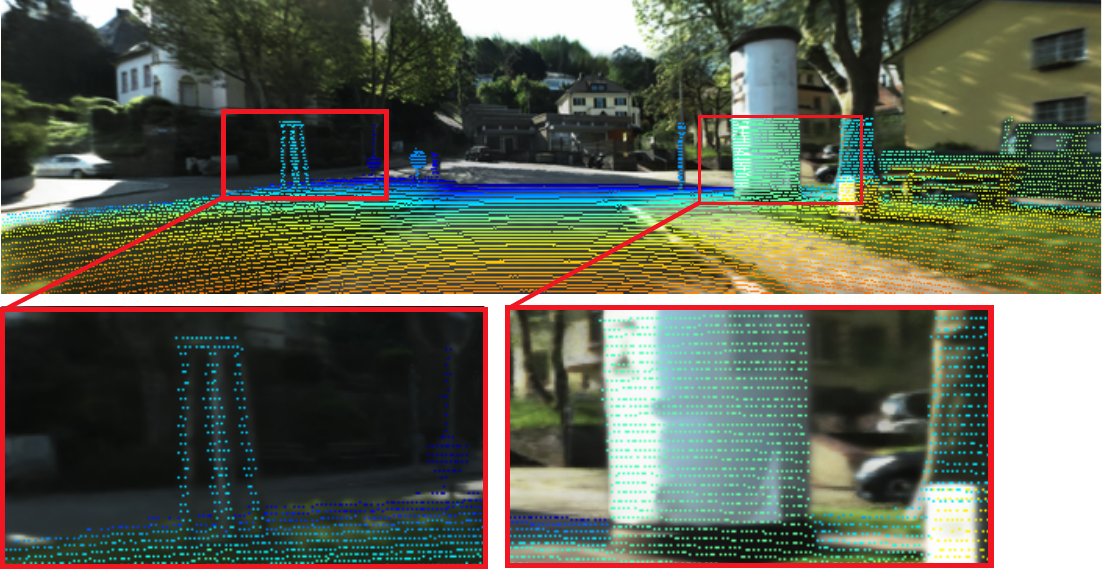}
    \end{subfigure}

     \caption{Limitation on KITTI-360 LiDAR ground truth calibration: we compare the alignment of re-projected 3D points from the LiDAR on the front image using KITTI extrinsic calibration and our optimized extrinsic calibration. We also re-project the 3D points on a synthetic image generated with the same pose as the LiDAR on the vehicle, in order to avoid parallax effect.}
    \label{fig:pc_into_images}
\end{figure*}

\begin{center}
\begin{table*}[hbt!]
\centering
\scriptsize
\caption{\label{tab:spatiotemporal_calibration} Spatiotemporal calibration accuracy.}

\renewcommand{\arraystretch}{1.2}
\begin{tabular}{@{} l l lll l lll l lll l lll l @{}}
\toprule
& \multicolumn{3}{c}{Front-Right Camera} && \multicolumn{3}{c}{Left-side Camera} \\ 
\cmidrule{2-4} \cmidrule{6-8} 
Sequence & Translation error (cm) & Rotation error (°) & Temporal error (ms) && Translation error (cm) & Rotation error (°) & Temporal error (ms) \\
\hline
0 & $1.1\pm0.3$ & $0.08\pm0.02$ & $0.5\pm0.3$ && $10.0\pm0.7$ & $0.41\pm0.24$ & $0.5\pm0.2$\\
1 & $2.0\pm0.3$ & $0.09\pm0.02$ & $0.9\pm0.3$ && $1.6\pm0.6$ & $0.16\pm0.03$ & $0.5\pm0.3$\\
2 & $1.8\pm1.0$ & $0.09\pm0.05$ & $0.5\pm0.4$ && $3.0\pm0.6$ & $0.06\pm0.05$ & $2.1\pm0.3$\\
4 & $1.9\pm0.7$ & $0.05\pm0.01$ & $1.3\pm0.6$ && $2.7\pm0.4$ & $0.05\pm0.25$ & $0.6\pm0.5$\\
& \multicolumn{3}{c}{Right-side Camera} && \multicolumn{3}{c}{LiDAR} \\
\hline
0 & $5.2\pm0.7$ & $0.12\pm0.02$ & $1.4\pm0.4$ && $9.3\pm2.3$ & $0.50\pm0.17$ & $0.04\pm0.02$\\
1 & $8.4\pm0.7$ & $0.26\pm0.02$ & $1.5\pm0.4$ && $10.6\pm3.0$ & $0.59\pm0.08$ & $6.9\pm2.1$\\
2 & $3.0\pm0.9$ & $0.07\pm0.05$ & $1.5\pm0.6$ && $15.8\pm2.9$ & $0.23\pm0.11$ & $2.3\pm1.7$\\
4 & $7.1\pm0.6$ & $0.08\pm0.03$ & $6.3\pm0.8$ && $14.0\pm2.1$ & $0.37\pm0.09$ & $22.4\pm1.8$\\
\bottomrule
\end{tabular}
\end{table*}
\end{center}
\begin{center}
\begin{table*}[hbt!]
\centering
\scriptsize
\caption{\label{tab:spatial_ablation} Calibration accuracy with solely rotation or translation error.}

\renewcommand{\arraystretch}{1.2}
\begin{tabular}{@{} l l ll l ll l l @{}}
\toprule
&& \multicolumn{2}{c}{Front-Right Camera} && \multicolumn{2}{c}{LiDAR} \\
\cmidrule{3-4} \cmidrule{6-7}
Error type & Initial error & Translation error (cm) & Rotation error (°) && Translation error (cm) & Rotation error (°)  \\
\hline
\multirow{3}{*}{Rotation all axes}
& 2° & $1.6\pm0.3$ & $0.09\pm0.01$ && $9.2\pm1.9$ & $0.39\pm0.09$ \\
& 5° & $1.7\pm0.5$ & $0.07\pm0.05$ && $9.9\pm2.3$ & $0.4\pm0.14$  \\
& 10° & $127.2\pm0.6$ & $15.81\pm0.02$ && $104.4\pm1.7$ & $17.07\pm0.13$  \\ 
\hline
\multirow{3}{*}{Translation all axes}
& 20cm & $1.8\pm0.4$ & $0.09\pm0.02$ && $8.7\pm1.5$ & $0.46\pm0.1$ \\
& 50cm & $1.7\pm0.5$ & $0.09\pm0.02$ && $7.8\pm1.2$ & $0.5\pm0.06$  \\
& 100cm & $1.7\pm0.3$ & $0.09\pm0.02$ && $8.8\pm2.4$ & $0.43\pm0.08$  \\ 
\bottomrule
\end{tabular}
\end{table*}
\end{center}
\subsection{Combined Calibration}\label{Combined Calibration}
In this section, we consider the full initial error as described in section~\ref{subsec:imp_detail}.
We run MOISST on all four cameras and the LiDAR of the KITTI-360 dataset.
We can see in Table~\ref{tab:spatiotemporal_calibration} that our method is able to calibrate the 2 side cameras, looking in completely different directions than our reference sensor.
The performance is variable depending on the sequence. For example, the sequence 0 is captured in a very narrow road, giving the side cameras a small FOV, reducing the overlap and causing a drop in accuracy.

\subsection{Discussions}
\subsubsection{Comparison with structure-based methods}
We wanted to compare our camera-LiDAR calibration results with the structure-based method from Yuan \textit{et al.}~~\cite{yuan2021pixel}, but we could only obtain subpar results with their method on the dataset we use (we obtained a mean translation error of 60 cm and 4.08° of rotational error starting from a translation and rotational error of 50 cm and 5°, respectively, on all axes). We found that it needed a denser point cloud from the LiDAR than what was provided in the KITTI-360 dataset in order to find reliable edge features in the scene. In addition, compared to our solution, this method is not able to do camera/camera and LiDAR/LiDAR calibration, or calibrate temporally.

\begin{center}
\begin{table*}[hbt!]
\centering
\scriptsize
\caption{\label{tab:opti_ablation} Ablation study on optimized parameters.}

\renewcommand{\arraystretch}{1.2}
\begin{tabular}{@{} l lll l lll l lll @{}}
\toprule
& \multicolumn{3}{c}{Front-Right Camera} && \multicolumn{3}{c}{LiDAR} \\
\cmidrule{2-4} \cmidrule{6-8}
Optimized parameters & Translation error (cm) & Rotation error (°) & Temporal error (ms) && Translation error (cm) & Rotation error (°) & Temporal error (ms) \\
\hline
Only spatial & $103.0\pm1.0$ & $1.03\pm0.03$ & -- && $104.8\pm2.3$ & $0.35\pm0.07$ & --\\
Only temporal & -- & -- & $89.6\pm0.2$ && -- & -- & $100.9\pm0.8$\\
Spatial \& temporal & $1.9\pm0.6$ & $0.08\pm0.01$ & $1.2\pm0.3$ && $11.5\pm2.1$ & $0.44\pm0.07$ & $10.1\pm1.3$\\
\bottomrule
\end{tabular}
\end{table*}
\end{center}
\begin{center}
\begin{table*}[hbt!]
\centering
\scriptsize
\caption{\label{tab:opti_pose_ablation} Poses accuracy for different optimized parameters.}

\renewcommand{\arraystretch}{1.2}
\begin{tabular}{@{} l lll l lll l lll @{}}
\toprule
& \multicolumn{2}{c}{Front-Right Camera} && \multicolumn{2}{c}{LiDAR} \\
\cmidrule{2-3} \cmidrule{5-6}
Optimized parameters & Translation error (cm) & Rotation error (°) && Translation error (cm) & Rotation error (°) \\
\hline
Only spatial & $9.2\pm1.2$ & $0.44\pm0.09$ && $12.6\pm1.7$ & $0.54\pm0.08$ & \\
Only temporal & $74.8\pm0.8$\ & $8.8\pm0.0$  && $90.4\pm1.1$ & $8.8\pm0.0$ & \\
Spatial \& temporal & $2.0\pm0.4$ & $0.09\pm0.02$ && $8.9\pm3.2$ & $0.42\pm0.09$ & \\
\bottomrule
\end{tabular}
\end{table*}
\end{center}
\subsubsection{Limitation of KITTI-360 LiDAR ground truth calibration}
\label{sec:limitation_kitti}
In our experiments, we found that the extrinsic calibration between the front camera and the LiDAR provided by KITTI-360 might be accurate only up to a few centimeters. To show this, we performed the following experiment: we re-projected the LiDAR points into the images captured by the front camera according to: 1) the provided ground truth calibration, 2) the extrinsic calibration we obtained after optimizing the spatiotemporal parameters. We also provided alignment comparison using NeRF generated images at the same location as the LiDAR position on the vehicle to avoid parallax effect. The LiDAR is positioned on top of the camera: some points re-projected on the images should not be visible from the camera position. Results are presented in Fig.~\ref{fig:pc_into_images} (more results in the supplementary video). Comparing the alignment between the re-projected 3D points on the real and synthetic images, we clearly see that our extrinsic calibration seems more accurate than the ground truth we use to compare our results with in this paper. 



\subsection{Ablation studies}
\label{sec:abla}
For the ablation studies, we only run our experiments on sequence 1.

\subsubsection{Rotation vs Translation error}
By running the training with solely rotation or translation errors of varying levels, we could observe that the initial rotation error has more impact on the final accuracy, as we did not get a satisfactory calibration when we introduced 10° rotation error on all axis. The results are shown in Table~\ref{tab:spatial_ablation}. On the contrary, the translation error is well-handled, even with 100 cm error set initially on all axis.

\subsubsection{Spatiotemporal coupling}
We run an ablation study on the optimized parameters and report the results in Table~\ref{tab:opti_ablation} and Table~\ref{tab:opti_pose_ablation}. 
It shows that if there is spatial and temporal errors and only one of them is optimized, it is not possible to obtain a correct calibration. Which means it is necessary to take into account both type of error. We can observe in Table~\ref{tab:opti_pose_ablation} that optimizing only the spatial parameters allows decent pose errors, showing that they are partly compensating the time offsets. This is possible because sequence 1 is mostly a straight line with the car driving at almost constant speed.

\begin{center}
\begin{table}[b]
\centering
\scriptsize
\caption{\label{tab:loss_ablation} Ablation study on losses with spatiotemporal calibration (\textbf{best} and \underline{second best}).}
\renewcommand{\arraystretch}{1.2}
\begin{tabular}{@{} l l ll l ll l l @{}}
\toprule
 & Translation error & Rotation error & Temporal error \\
Loss & (cm) & (°) &  (ms) \\
\hline
$L_C$ + $L_D$ & $7.4\pm1.2$ & $0.31\pm0.06$ & $4.4\pm1.0$ \\
$L_C$ + $L_D$ + SSIM & $\underline{6.9\pm1.3}$ & $\underline{0.30\pm0.04}$ & $\bm{1.9\pm0.6}$ \\
$L_C$ + $L_D$ + DS & $14.4\pm2.1$ & $0.31\pm0.14$ & $4.4\pm0.8$ \\
$L_C$ + $L_D$ + SSIM+DS & $\bm{6.8\pm1.0}$ & $\bm{0.28\pm0.06}$ & $\underline{2.8\pm0.6}$ \\
\bottomrule
\end{tabular}
\end{table}

\end{center}
\subsubsection{Ablation on additional losses}
In Table~\ref{tab:loss_ablation}, we demonstrate that the overall accuracy of our method increases when $\mathcal{L}_{SSIM}$ and $\mathcal{L}_{DS}$ are used. $\mathcal{L}_{SSIM}$ has the largest impact on the performance as it help the implicit scene representation to learn a proper and sharp geometry from radiometric signals. It makes sense that better scene geometry improves the calibration accuracy, especially between LiDARs and cameras.

\section{CONCLUSIONS AND FUTURE WORK}
We presented in this paper MOISST, a novel approach based on implicit neural scene representation to spatially and temporally calibrate a multi-sensor system. The proposed approach has the advantage of being scalable to any number of cameras and LiDARs by relying on the trajectory of a single reference sensor. The proposed approach does not require any targets, or specific geometric structure within the scene to achieve accurate results. It is fully automatic and relies on gradient descent to optimize the calibration parameters.
In the future, we expect to address some limitations of the method by calculating the poses of the reference sensor automatically instead of relying on the given ground truth, and by finding a way to bypass the need of priors for the other sensors.
We would also like to add larger compatibility to other types of sensor, such as rolling shutter cameras or distorted LiDARs, and the optimization of intrinsic parameters. 
Finally, we would implement the multi-scene optimization, which should improve robustness by relying on more varied scenes to optimize a specific multi-sensor system, as well as the ability to manage dynamic elements in the scene, which are not considered currently.

\addtolength{\textheight}{-6cm}   






{\small
\bibliographystyle{IEEEtran}
\bibliography{biblio}
}

\end{document}